\pdfoutput=1

\documentclass[11pt]{article}

\usepackage[final]{acl}

\usepackage{times}
\usepackage{latexsym}
\usepackage{multirow}
\usepackage{algorithm}
\usepackage{algpseudocode}
\usepackage{booktabs}
\usepackage{amsmath}
\usepackage{enumitem}
\usepackage{array}

\usepackage[T1]{fontenc}

\usepackage[utf8]{inputenc}

\usepackage{microtype}

\usepackage{inconsolata}
\usepackage{algorithm}
\usepackage{algorithmicx}
\usepackage{algpseudocode}
\usepackage{graphicx}
\usepackage[dvipsnames]{colortbl}
\usepackage{adjustbox}
\usepackage{subcaption}
\usepackage{xcolor}
\usepackage{easyReview}
\setreviewsoff

\usepackage{tcolorbox}
\definecolor{Gainsboro}{rgb}{0.86, 0.86, 0.86}
\definecolor{mygreen}{RGB}{153,255,51}

\newcommand{\modelname}{MMAPG}
\newcommand{\graphname}{Adaptive Planning Graph}

\title{\modelname{}: A Training-Free Framework for Multimodal Multi-hop Question Answering via \graphname{}s}

\author{
 \textbf{Yiheng Hu\textsuperscript{1,2}},
   \textbf{Xiaoyang Wang\textsuperscript{1}},
 \textbf{Qing Liu \textsuperscript{2}}, 
 \textbf{Xiwei Xu \textsuperscript{2}}, 
 \\
 \textbf{Qian Fu \textsuperscript{2}}, 
 \textbf{Wenjie Zhang\textsuperscript{1}},
 \textbf{Liming Zhu \textsuperscript{2}}, \\
  \textsuperscript{1}University of New South Wales, Sydney, Australia\\
 \textsuperscript{2}CSIRO Data61, Australia\\
 \texttt{\{yiheng.hu, xiaoyang.wang1, wenjie.zhang\}@unsw.edu.au}\\
 \texttt{\{Q.Liu, Xiwei.Xu, Qian.Fu, Liming.Zhu\}@data61.csiro.au}
 }

\begin{document}
\maketitle
\begin{abstract} 
Multimodal Multi-hop question answering requires integrating information from diverse sources, such as images and texts, to derive answers. Existing methods typically rely on sequential retrieval and reasoning, where each step builds on the previous output. However, this single-path paradigm makes them vulnerable to errors due to misleading intermediate steps. Moreover, developing multimodal models can be computationally expensive, often requiring extensive training. To address these limitations, we propose a training-free framework guided by an \graphname{}, which consists of planning, retrieval and reasoning modules. The planning module analyzes the current state of the \graphname{}, determines the next action and where to expand the graph, which enables dynamic and flexible exploration of reasoning paths. To handle retrieval of text to unspecified target modalities, we devise modality-specific strategies that dynamically adapt to distinct data types. Our approach preserves the characteristics of multimodal information without costly task-specific training, enabling seamless integration with up-to-date models. Finally, the experiments on MultimodalQA and WebQA show that our approach matches or outperforms existing models that rely on training.
\end{abstract}

\section{Introduction}

\begin{figure}[h]
  \centering
  \includegraphics[width=0.9\linewidth]{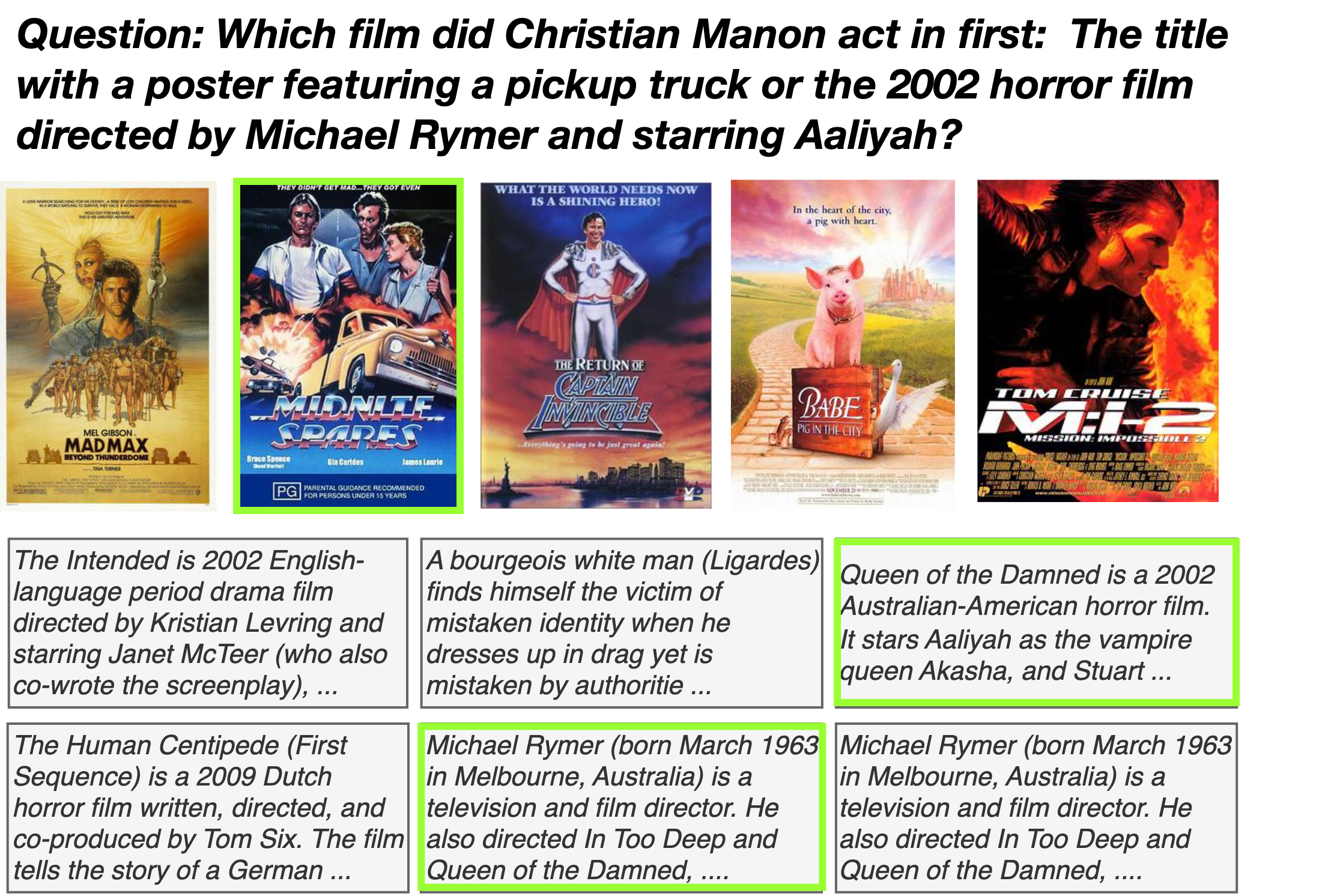}
  \caption{An example of multimodal multi-hop QA. It requires identifying relevant information (bounded by \textcolor{mygreen}{\textbf{\textit{green box}}}) from diverse sources to generate answers.}
  \label{fig.complexqa_example}
\end{figure}
The field of question answering(QA) has gained significant attention and is increasingly applied across various domains, including customer support, healthcare, and education, particularly with the rapid advancements driven by large language models (LLMs)~\cite{su2019generalizing, lu2022learn, wei2022chain, shao2023prompting, he2024g}. These models have demonstrated strong performance in single-hop QA.
However, multimodal multi-hop QA~\cite{talmor2021MultimodalQA, chang2022webqa} presents a greater challenge, as it requires integrating diverse sources. As illustrated in Figure~\ref{fig.complexqa_example}, relevant information must be identified across multiple sources with different modalities to generate answers.  
In these settings, only a subset of sources is relevant, while others introduce noise. 
Solving this task requires approaches that effectively integrate both retrieval and reasoning capabilities. 

\begin{figure*}[h]
  \centering
  \includegraphics[width=0.9\linewidth]{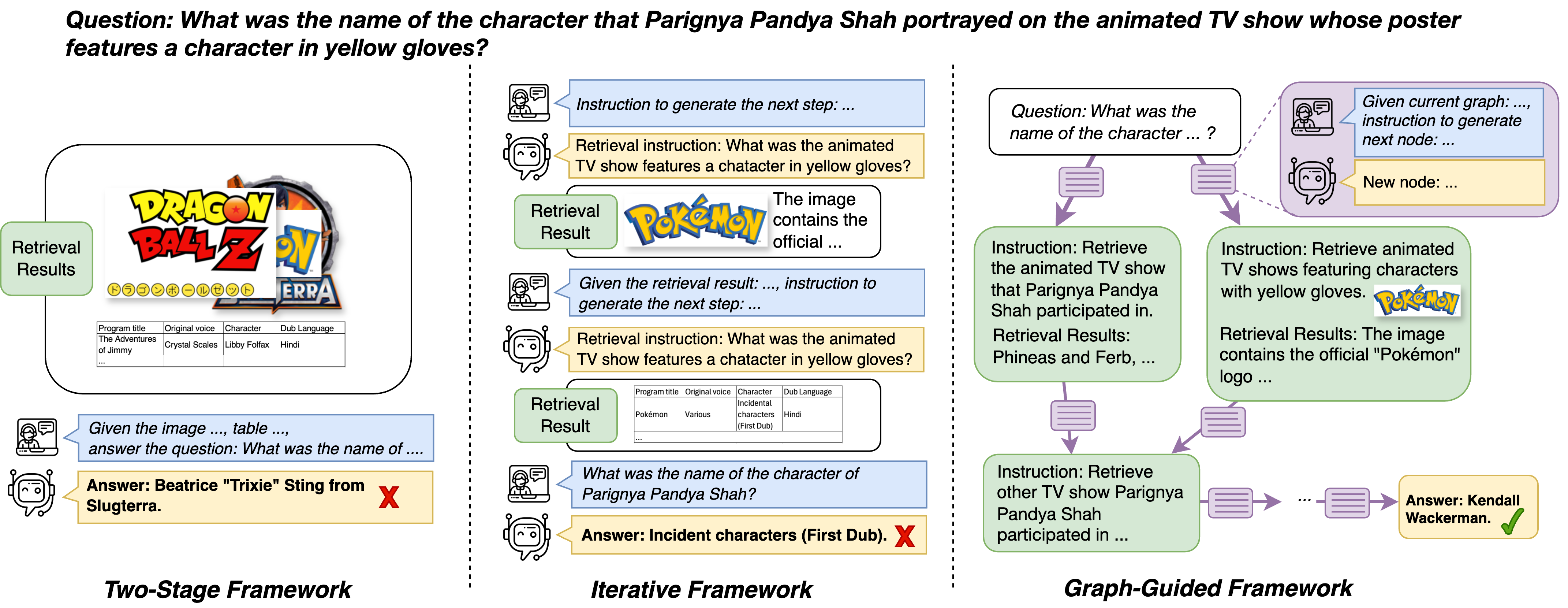}
  \caption{A representative case illustrating how each framework handles deviations in the reasoning process to reach the correct answer. We prompt the LLM to operate following the paradigm of each framework, comparing how structural differences affect reasoning under the same conditions.
}
  \label{fig:compare_main_approach}
\end{figure*}

Current research in this area centers around two main paradigms. 
The first approach adopts a two-stage framework~\cite{yu2023unified, liu2023mmhqa, lim2024unirag}. It retrieves all potentially relevant information in a single step, followed by answer generation based on the retrieved information. The two stages are trained independently with distinct objectives, which can lead to a misalignment. Specifically, the reasoning stage implicitly assumes that the retrieved sources are complete and accurate, introducing fragility as early retrieval failures cannot be rectified during generation. 
For example, in Figure~\ref{fig:compare_main_approach}, the retrieval results linked to \textit{animated TV show} and \textit{yellow gloves} involve multiple false positives, and overlook the correct poster. In the next stage, the system is forced to reason from the incorrect contexts, leading to inaccurate answer.

The second mainstream method employs an iterative approach~\cite{trivedi2022interleaving, yang2023enhancing, zhang2024entailment},
which offers greater flexibility as it has no restrictions on the number of steps, allowing for the integration of additional sources as needed. 
The structure of these methods follow a single-path paradigm, where actions proceed in a fixed pattern.
\add{This design introduces cascading error propagation, which relies on the model's inherent reasoning ability to correctify.}
For instance, in Figure~\ref{fig:compare_main_approach}, if an initial retrieval incorrectly identifies details about the TV show \textit{Pokémon}, this mistake could mislead the next subquestion, ultimately leading to an incorrect result along a single-path with flaw.
The challenge is further amplified in multimodal settings, where the modality involved in each step is often unknown. 
To handle different types of sources, SKURG~\cite{yang2023enhancing} fuses image and text embeddings to create an entity-centered representation, which relies on extensive training of existing models.
ETG~\cite{zhang2024entailment} retrieves evidences by converting images into texts during preprocessing, which may result in information loss.
and omit relevant contexts.
These limitations highlight the need for developing methods that can dynamically adapt to multimodal sources. 

To address these issues, we introduce a training-free \graphname{}-guided approach.
As illustrated in Figure~\ref{fig:compare_main_approach}, both the two-stage and iterative framework rely on a single-path paradigm, where each step strictly depends on the previous one. Consequently, errors in earlier steps can propagate and affect the final answers. 
In our framework, we adopt a \graphname{}, where each node represents a thought or a result generated by module, and edges denote the dependencies between nodes.
It presents a more flexible flow, allowing the new steps to be dynamically generated from any relevant node in the \graphname{}. 
The proposed method consists of planning, retrieval and reasoning modules.
At each step, The planning module generates a plan for one action on-the-fly. It continuously analyzing the current state to determine the next move. 
To facilitate search of contexts across multiple modalities, existing works either convert images into texts which may result in incomplete captions, or rely on resource-intensive pretraining.  
To address these limitations, our module constructs separate knowledge bases and uses modality-specific strategies to collect relevant information. It enables effective retrieval while preserving modality details and avoiding additional computational costs.
The main contributions of our paper are presented as follows: 
\begin{enumerate}
    \item We introduce a novel framework, \modelname{} (\underline{M}ultimodal \underline{M}ulti-hop \underline{A}daptive \underline{P}lanning \underline{G}raph), that offers enhanced flexibility for tackling multimodal multi-hop QA. By constructing the \graphname{} step-by-step, our approach facilitates dynamic exploration of different sources 
    and supports a graph-based reasoning flow.
    \add{To the best of our knowledge, this is the first work using graph-based planning for multimodal multi-hop QA.}
    \item We address multimodality with a training-free framework by employing distinct off-the-shelf within specialized modules. 
    Our proposed modules preserve the details of each modality while leveraging the generated rationale to support the construction of the \graphname{}.
    \item We conduct experiments on MultimodalQA and WebQA datasets. The results demonstrate that our model performs comparably or better than trained models.
\end{enumerate}

\section{Related Works}

\subsection{Multimodal Multi-hop QA}
The first mainstream approach to solve multimodal multi-hop QA is the two-stage framework. 
For instance, Solar~\cite{yu2023unified} and UniRAG~\cite{lim2024unirag} both unify multimodal sources into texts, retrieve top-$k$ results and employ language models to generate the final response. 
To handle multimodality,  AutoRouting and ImplicitDecomp~\cite{talmor2021MultimodalQA} fine-tune models to answer questions depending on modality identified by a classifier.
Meanwhile, PERQA~\cite{yang2023progressive} employs an iterative evidence selection process and incorporates multimodal reasoning during the generation phase. 
In contrast to the methods previously discussed, which all involve training or fine-tuning, MMHQA-ICL~\cite{liu2023mmhqa} represents a training-free paradigm, which autonomously generates prompts for in-context learning.

The second approach focuses on an iterative framework, which is widely used in single-modality QA~\cite{trivedi2022interleaving}. 
However, its application in multimodal multi-hop QA is less common. 
A notable example is SKURG~\cite{yang2023enhancing}, which introduces a unified retrieval and generation module that iteratively integrates multimodal information.
A recent work, ETG~\cite{zhang2024entailment}, proposes a mixture-of-experts approach to combine retrieval and generation, with reasoning represented through an entailment tree.

\subsection{Chain of thought}
The chain of thought (CoT) reasoning has significantly improved LLMs' reasoning abilities. It inspires approaches that prompt LLMs to generate full reasoning at once~\cite{kojima2022large} or incrementally~\cite{xu2023rewoo,shen2024hugginggpt}. While these methods follow a single-path paradigm, multi-path approaches like CoT-SC~\cite{wang2022self}, Tree-of-Thought (ToT)~\cite{yao2024tree}, and Graph-of-Thought (GoT)~\cite{besta2024graph} explore multiple reasoning paths and decision-making processes. Notably, GoT requires users to manually define the execution plan, making it less flexible for QA tasks that require adaptive plans for different questions. 

Despite these advances, these reasoning methods might overlook the integration of external knowledge sources into the reasoning process. To address this limitation, recent research has proposed incorporating diverse knowledge sources into reasoning pipelines~\cite{tan2025paths, sun2023think, sarmah2024hybridrag, tan2025hydra, su2023semi, ma2023chain}.

\begin{figure*}[h]
  \centering
  \includegraphics[width=0.9\linewidth]{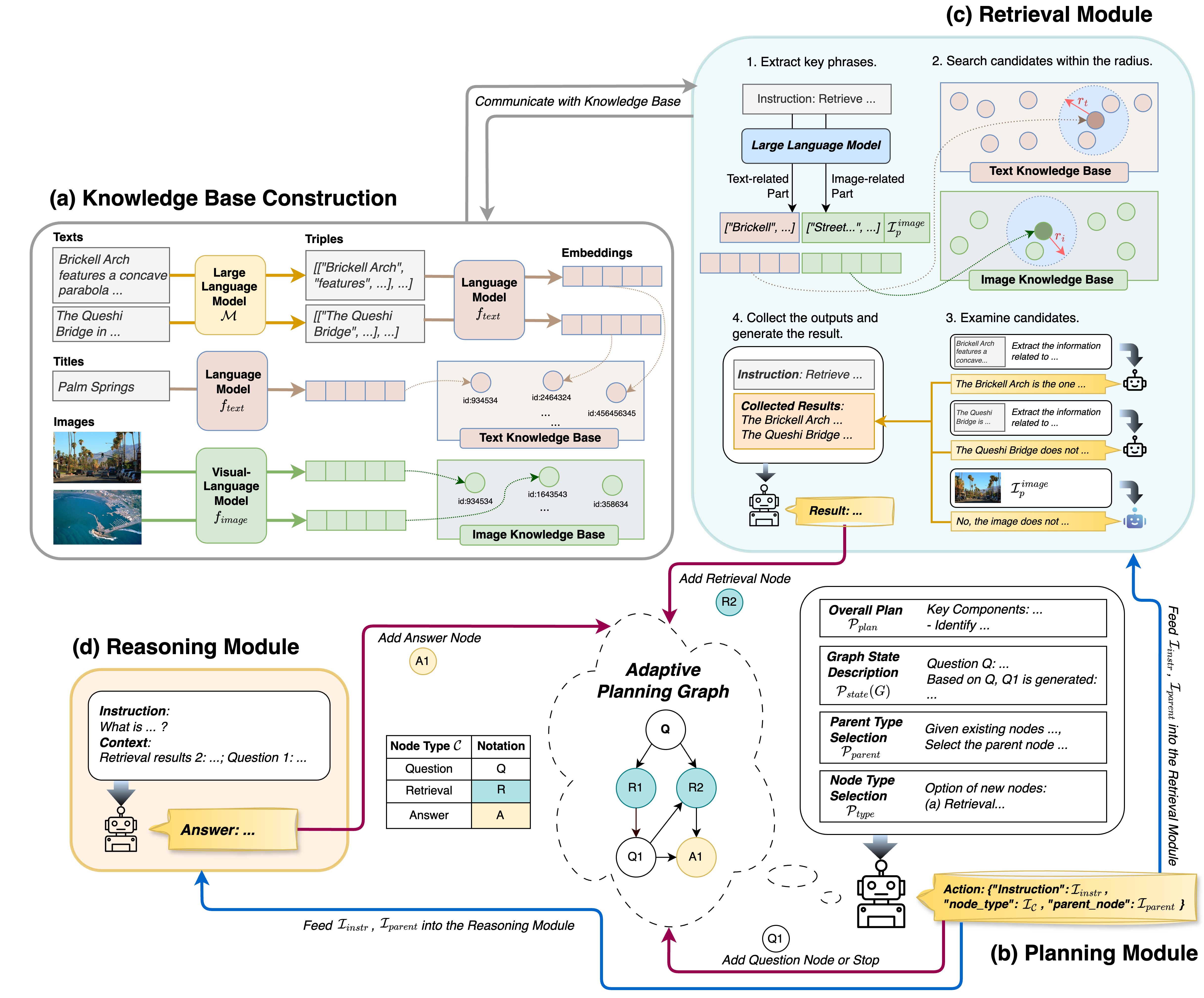}
  \caption{An overview of our framework \modelname{}, which consists of four parts: \textbf{(a) Knowledge Base Construction}: precomputes embeddings for given sources; \textbf{(b) Planning Module}: determines the next action in the \graphname{} and which module to call; \textbf{(c) Retrieval Module}: retrieve relevant sources and find key information; \textbf{(d) Reasoning Module}: derives the answer based on the provided instruction. }
  \label{fig:framework}
\end{figure*}

\subsection{Multimodal Retrieval}
Retrieval across various modalities has been extensively researched, usually with fixed source and target modalities, such as text-to-image, image-to-text, or image-text pair to image retrieval. However, retrieval without predefined target modalities has received less attention.
Previous research~\cite{mayilvahanan2023does} has demonstrated that similarity scores between intra-modalities and inter-modalities exhibit different distributions, presenting inherent challenges in this area. MuRAG~\cite{chen2022murag} pretrains a multimodal retrieval model, but requires collecting a large number of samples. REVEAL~\cite{hu2023reveal} involves pretraining and developing gating score to select dataset. These approaches illustrate that facilitating multi-modal retrieval often relies on costly resources.
Other works~\cite{yu2023unified, liu2023mmhqa} attempts to convert all the images into texts to address the challenges of multimodal retrieval. However, questions can focus on various details within an image, and these critical information may not be preserved when converting to texts.

\section{Methods}
We present our motivation for graph-guided framework in Section~\ref{sec:motivation}. Then we depict the main modules in our framework \modelname{}, including knowledge base construction (Section~\ref{sec:kb_construction}), planning module (Section~\ref{sec:planning}) and retrieval module (Section~\ref{sec:retrieval}). The overall workflow is detailed in Appendix~\ref{appd:alg}. 
Since the reasoning module is simply implemented by calling an off-the-shelf model to generate the answer, we omit its  discussion here, where its details can be found in Figure~\ref{fig:framework}.

\subsection{\modelname{} Overview} \label{sec:motivation}
In this section, we demonstrate how our graph-guided framework alleviates the limitations of two-stage and iterative framework based methods.
To establish the framework, we prompt the LLM to analyze the current graph and generate instructions for next steps, as shown in Figure~\ref{fig:compare_main_approach}.
Each step in the reasoning process is represented as a node, and we explicitly allow the new nodes to be created based on any existing node.
This capability ensures that even if one path proves ineffective, the framework can still identify alternative paths based on the current nodes. In the example, when the retrieval of animated TV shows featuring two yellow gloves is unsuccessful, the system can combine existing nodes to explore other paths for retrieving relevant TV shows within a limited range.
Compared to previous approaches, our paradigm enhances the flexibility and allows for a more adaptive process.

\subsection{Knowledge Based Construction} \label{sec:kb_construction}
Given a set of sources $S = \{ S_1, \cdots, S_n \}$.
Each source $S_i$ may have associated text-based components $T_i$, (e.g., $T_i = \{ T_i^{title}, T_i^{caption} \}$).
We maintain separate knowledge base for text and images, denoted as $KB_{text}$ and $KB_{img}$.

Specifically, the text knowledge space consists of embedding derived from textual sources and text-based image information, including image titles and image captions.
Since different types of text-based information vary in length and contextual detail, which can impact the retrieval performance~\cite{wang2024searching}, to ensure consistent granularity, we decompose $S_i$ into relational triplets $\mathcal{T}_i = \{ \tau_{i1} \cdots, \tau_{i m_i} \}$.
Here, we employ a few-shot prompting method~\cite{zhang2024extract} to extract triplets from $S_i$, and compute the embedding: 
\begin{align}
    &\mathcal{T}_i  = \mathcal{M} (\mathcal{P}_{tri} \oplus \mathcal{F}_{tri} \oplus S_i ), \\
    & \mathbf{e}_{ij}^{text} = f_{text} (\tau_{ij} ),
\end{align}
where $\mathcal{M}$ represents the LLM used; $\mathcal{P}_{tri}$ is the prompt; $\mathcal{F}_{tri}$ is the few-shot examples; $\oplus$ means concatenation; $f_{text}$ is the model to generate text embedding; $\mathbf{e}_{ij}^{text}$ is the computed embedding for triplet $\tau_{ij}$. For short components, such as titles, we directly compute their embeddings. Finally, we construct the text knowledge base $KB_{text}$ by storing all the text embeddings $\{\mathbf{e}_1^{text},\cdots,\mathbf{e}_N^{text}\}$.

To construct image knowledge base $KB_{img}$, embeddings are directly computed as follows: 
\begin{align}
    \mathbf{e}_i^{img} = f_{img} (S_i), 
\end{align}
where $f_{img}$ is the multimodal model to generate image embedding. Similarly, $KB_{img}$ stores all the image embeddings $\{\mathbf{e}_1^{img},\cdots,\mathbf{e}_M^{img}\}$.

\subsection{Planning Module} \label{sec:planning}
In this section, we introduce the planning module, the core component for constructing the \graphname{}, denoted as $G = (V, E)$. The node $v \in V$ represents the action at each stage, and directed edges $e \in E$ represent dependencies between nodes. Specifically, $e = (v_i, v_j)$ indicates that node $v_j$ is built upon $v_i$, which reflects the logical flow of global reasoning. 
The \graphname{} is built by adding new nodes along with their corresponding edges on the fly, where the generation of these nodes and edges are informed by the planning module at each step.

To harness the reasoning capability of LLMs for global planning, we design the prompt that serves two functions: analyzing the current state of the \graphname{} as it evolves to answer the question and determining the next appropriate action. For the first function, we provide both an overall plan outlining the essential information needed and a summary of the current graph that reflects the progress made so far. Together, they offer the model a clear perspective of the information already gathered and what remains to be explored. For the second function, we present a set of options for expanding the graph. It allows the model to decide the most appropriate node to generate based on its prior analysis. 
To facilitate these processes, our prompts consists of four components: 
(1) Overall plan, denoted as $\mathcal{P}_{plan}$. Given that many LLMs exhibit limitations in handling long-term planning, we generate a high-level guide, which outlines the key components and possible global plans. It serves as a reference for expanding the \graphname{} $G$. 
(2) Graph State Description, presented as $\mathcal{P}_{state}(G)$. It displays the current state of the \graphname{} by describing the content of existing nodes and their dependencies. 
(3) Parent Node Selection Instruction, represented as $\mathcal{P}_{parent}$. It instructs the system to select parent nodes from all nodes in the graph $G$. 
(4) Node Type Selection Instruction, denoted as $\mathcal{P}_{\mathcal{C}}$. It provides a set of node types to select. 
We concatenate these components to build prompts that are fed into LLM $\mathcal{M}$: 
\begin{align}
\begin{split}
\mathcal{I}_{\mathcal{C}}, \mathcal{I}_{parent} , \mathcal{I}_{instr}  &= \mathcal{M} (\mathcal{P}_{plan} \oplus \mathcal{P}_{state}(G)  \\
& \oplus \mathcal{P}_{parent} \oplus \mathcal{P}_{\mathcal{C}}), 
\end{split} 
\end{align}
where $\mathcal{I}_{\mathcal{C}}, \mathcal{I}_{parent} , \mathcal{I}_{instr}$ are the type, parent nodes, and instruction for generating the content of the new node. 
The type of new node $\mathcal{C}_i \in \{ Question, Answer, Retrieval, Stop\}$ is informed by $\mathcal{I}_{\mathcal{C}}$. It signifies the system of the next action to take. 
We present the instructions and actions according to each node type as follows:
\begin{itemize}
    \item {\textbf{Question}}: The instruction is a direct question. Since no further processing is required, this question is taken as the new node content.
    \item {\textbf{Answer}}: The instruction specifies which question to be answered. The instruction and parent nodes are passed into the reasoning module to generate the corresponding answer. The answer is then added as the new node content. 
    \item {\textbf{Retrieval}}: The instruction outlines the information to be retrieved. 
    The instruction and parent nodes are passed to the retrieval module. If relevant information is found, it is added as the new node content. However, if no relevant information is retrieved, a node is created to indicate that no results were found. 
    \item {\textbf{Stop}}: 
    The action with this node is to terminate the process. It uses the content of the last answer node as the final answer. If an answer has not yet been generated, the LLM will generate one based on the overall graph.
\end{itemize}
The graph is then expanded by adding the new node along with the corresponding edges, which are determined by $\mathcal{I}_{parent}$. For instance, if $\mathcal{I}_{parent} = \{ v_j\}$, a directed edge $e_{ij}= \{v_j, v_i \}$ will be added to the \graphname{}, to indicate $v_i$ is derived from the thought of $v_j$.
The planning module will continuously plan the next step to update the graph until a stop node is generated or the maximum number of iteration is reached.

\subsection{Retrieval Module} \label{sec:retrieval}
The retrieval module is responsible for extracting relevant information based on the given instruction and parent nodes.
To eliminate training costs and enhance generalizability, we leverage off-the-shelf models to handle multimodality. However, retrieval across different source types \add{can exhibit inherently different similarity score distributions, as we show in detail in Appendix~\ref{appd:retrieval_challenges}.} 
To address this, we propose tailored strategies for different modalities.
We firstly utilizes LLM to decompose the instruction into text-related and image-related components: 
\begin{align}
    \mathcal{I}_{instr}^{text}, \mathcal{I}_{instr}^{img} = \mathcal{M} (\mathcal{P}_{decomp} \oplus \mathcal{I}_{parent} \oplus \mathcal{I}_{instr}),
\end{align}
where $\mathcal{P}_{decomp}$ is the prompt for decomposition. 
Next, we apply different methods to extract key elements from text-related and image-related parts.

For the text-related part, we employ few-shot examples mined from the dataset to identify key phrase. The queries utilized for text retrieval are generated as follows: 
\begin{align}
    \mathcal{Q}_{text} = \mathcal{M} (\mathcal{P}_{extract} \oplus \mathcal{F}_{text}\oplus \mathcal{I}_{instr}^{text}),
\end{align}
where $\mathcal{P}_{extract}$ is the prompt for key phrase extraction, and $\mathcal{F}_{text}$ represents the set of few-shot examples to identify key phrases or words.

For the image-related part, the type of image retrieval is identified from $\mathcal{I}_{instr}^{img}$ firstly, which determines the action to be taken next. There are two types of image retrieval defined.
The first one is targeted image retrieval. It occurs when the instruction $\mathcal{I}_{instr}^{text}$ mentions specific identifiers of a particular image, such as title. 
We expect these identifiers to be extracted and used to search the text knowledge base $KB_{text}$. The query $\mathcal{Q}_{text}'$ is generated by LLM as follows: 
\begin{align}
& \mathcal{Q}_{text}', \mathcal{I}_{tgt}^{img} = \mathcal{M} (\mathcal{P}_{tgt}  \oplus \mathcal{F}_{tgt} \oplus \mathcal{I}_{instr}^{img}),  
\end{align}
where $\mathcal{P}_{tgt}$ and $\mathcal{F}_{targe}$ are the prompt and few-shot examples for targeted image retrieval. The generated $\mathcal{I}_{tgt}^{img}$ will be used in later steps for candidate examination. The final query to search in text knowledge base $KB_{text}$ is then the combination of $\mathcal{Q}_{text}$ and $\mathcal{Q}_{text}'$: 
\begin{align}
    \mathcal{Q}_{text} = \mathcal{Q}_{text} \oplus \mathcal{Q}_{text}'.
\end{align}

The other type is descriptive image retrieval, which is utilized when a description about the image content is provided. In this case, the image-related part instruction $\mathcal{I}_{instr}^{img}$ usually contains descriptive text to guide the system in locating images that best match the description in $KB_{img}$. The queries are generated as follows:
\begin{align}
    \mathcal{Q}_{img}, \mathcal{I}_{descr}^{img} = \mathcal{M} (\mathcal{P}_{descr}  \oplus \mathcal{F}_{descr} \oplus \mathcal{I}_{instr}^{img}), 
\end{align}
where $\mathcal{P}_{descr}$, $\mathcal{F}_{descr}$ are the prompt and few-shot examples for descriptive image retrieval.

Next, we generate the corresponding embeddings for extracted queries. These embeddings are then searched in the corresponding knowledge base to identify matches within a defined radius $r_t$ and $r_i$ for $KB_{text}$ and $KB_{img}$ respectively. For example, when searching within the text knowledge base, we first compute the embedding for the extracted phrase $q_i^{text}$ as follows: 
\begin{align}
    \mathbf{e}_{i}^{text} &= f_{text} (q^{text}_i).
\end{align}
Next, we identify the candidates from $KB_{text}$ that are within the defined radius $r_t$ from $\mathbf{e}_{i}^{text}$:
\begin{align}
    C^{text} &= \{ \mathbf{e}_j \in KB_{text} \mid \|\mathbf{e}_j - \mathbf{e}_{i}^{text} \| \leq r_t \}.
\end{align}
Then, we could derive $k$ candidates as $C^{text}= \{ c_1^{text}, \cdots, c_k^{text}  \}$.

\begin{table*}[htbp]
\centering
\begin{adjustbox}{width=1.9\columnwidth,center}
\setlength\tabcolsep{5pt}
  \begin{tabular}{@{} l p{7.0cm} >{\centering\arraybackslash}p{4.7cm}  cc cc cc @{}}
    \toprule
 &  {\multirow{2}{*}{\textbf{Methods} }} &  \multirow{2}{*}{Trained Models} & \multicolumn{2}{c }{\textbf{Single-Modal}} & \multicolumn{2}{ c}{\textbf{Multi-Modal}} & \multicolumn{2}{ c}{\textbf{Overall}} \\
     \cmidrule(lr){4-5}\cmidrule(lr){6-7}\cmidrule(lr){8-9}
 &  &  & F1 & EM & F1 & EM  & F1 & EM  \\
    \hline
    \rowcolor{Gainsboro}\multicolumn{9}{l}{\textit{Fine-tuned}} \\
    &\quad AutoRouting \cite{talmor2021MultimodalQA} & \textit{RoBERTa, ViLBERT} & 58.5 & 51.7 & 40.2 & 34.2 & 51.1 & 44.7 \\
    & \quad ImplicitDecomp \cite{talmor2021MultimodalQA} & \textit{RoBERTa, ViLBERT} & 58.8 & 51.1 & 51.7 & 46.5 & 55.9 & 49.3\\
   &  \quad Solar \cite{yu2023unified} &\textit{BERT, T5} & 74.8 & 69.7 & 65.4 & 55.5 & 66.1 & 59.8\\
    &  \quad  PERQA \cite{yang2023progressive} &\textit{BERT, ViT+Llama+Lora} & 74.1 & 69.7 & 60.3 &  54.7 & 67.8 & 62.8   \\ 
 & \quad  SKURG \cite{yang2023enhancing} & \textit{OFA,BART} & 69.7 & 66.1 & 57.2 & 52.5 & 64.0 & 59.8\\
 &   \quad  ETG \cite{zhang2024entailment} & \textit{T5(MoE)} &74.9  & \textbf{69.8} &\textbf{ 65.7} & \textbf{64.7} & 66.5 & \textbf{68.2}\\
    \hline
    \rowcolor{Gainsboro}\multicolumn{9}{l}{\textit{W/o fine-tuning}}\\
  &   \quad MMHQA-ICL \cite{liu2023mmhqa}& \textit{-} & 72.9  & 60.5 & 55.5  & 46.2 & 65.8 & 54.8  \\
   & \quad  \modelname{} (ours)  & \textit{-} & \textbf{75.4} & 65.2 & 65.0 & 51.9  &  \textbf{70.6 }& 59.1 \\ \bottomrule
\end{tabular}
\end{adjustbox}
  \caption{The comparison of different methods on MultimodalQA dataset. We report the F1 and EM scores for single-modal, multi-modal and overall questions. The trained models used for each model are also presented.}
  \label{tab:mmqa_modality_result}
\end{table*}

While retrieval based solely on similarity score may introduce many outliers, a more refined examination is conducted using off-the-shelf models. These models assist in verifying whether the content of the candidates is relevant to the instruction. For text-related parts, we directly instruct LLM to extract useful information $O^{text}_i$ from $c_i^{text}$: 
\begin{align}
    O^{text}_i = \mathcal{M} (\mathcal{P}_{exam}^{text} \oplus \mathcal{I}^{text}_{instr} \oplus c_i^{text}),
\end{align}
where $\mathcal{P}_{exam}^{text}$ is the prompt used for examining textual candidates.
For image-related part, we instruct vision-language model $\mathcal{M}_{vl}$ to examine the image: 
\begin{align}
    O^{img}_j = \mathcal{M}_{vl} (\mathcal{P}_{exam}^{img} \oplus \mathcal{I}_{p}^{img} \oplus c_j^{img}).
\end{align}
where $p\in \{descr,target\}$.
Finally, all the examination results $O = \{O_1^{text}, \cdots, O_1^{img}, \cdots \}$ will be collected together. These compiled information, along with $\mathcal{I}_{instr}$, is fed into the LLM for information extraction: 
\begin{align}
    R = \mathcal{M} (P_{retr} \oplus \mathcal{I}_{instr} \oplus O).
\end{align}

The final results $R$ are then served as the content of the new retrieval node.

\section{Experiments}

\subsection{Experimental Setup}
We utilize MultimodalQA~\cite{talmor2021MultimodalQA} and WebQA~\cite{chang2022webqa} datasets for multimodal multi-hop QA evaluation. 
MultimodalQA is a dataset designed for question answering across text, tables, and images, where each question is accompanied by a set of distractors. The answers are evaluated using  the F1 and exact match (EM) scores. 
WebQA consists of QA pairs along with images or text snippets, including distractors. The evaluation metrics for WebQA involve QA-Acc, keyword-based accuracy and QA-FL, which assesses fluency using BARTScore.
The implementation details of our framework can be found in Appendix~\ref{appd:impl}.

\begin{table}[htbp]
\begin{adjustbox}{width=.7\columnwidth,center}
\setlength\tabcolsep{7pt}
\centering
  \begin{tabular}{@{} p{3.5cm}  cc @{}}
    \toprule
    \textbf{Methods} & QA-Acc & QA-FL\\
    \hline
    \rowcolor{Gainsboro}\multicolumn{3}{l}{\textit{Fine-tuned}} \\
    \quad  Solar & 58.9 & 60.9 \\ 
    \quad MuRAG & 54.6 & 55.7 \\
   \quad  SKURG & 63.4 & 47.8\\
   \quad  PERQA &  63.9  & \textbf{61.7}  \\
   \hline
    \rowcolor{Gainsboro}\multicolumn{3}{l}{\textit{W/o fine-tuning}} \\
   \quad  \modelname{} (ours) &  \textbf{65.9} & 56.4 \\
  \bottomrule
\end{tabular}
\end{adjustbox}
  \caption{The performance comparison on WebQA.}
  \label{tab:webqaoverall}
\end{table}

\subsection{Main Results}
We present our results on the MultimodalQA in Table~\ref{tab:mmqa_modality_result}. Compared to baseline models, our method achieves the highest F1 scores in the overall evaluation. However, our exact match scores fall short of the current state-of-the-art. This disparity arises since our approach operates without fine-tuning, and thus, does not align precisely with the ground-truth labels provided by the dataset (See Appendix~\ref{appd:err} for more details). 
Our single-modality F1 score surpasses existing methods, although the multimodal F1 result remains slightly below that of fine-tuned models. 

\begin{figure}
  \centering
  \includegraphics[width=0.8\linewidth]{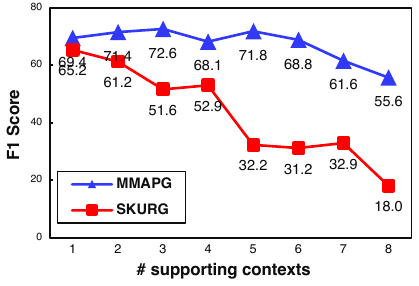}
  \caption{The F1 scores of our method and SKURG across different numbers of supporting documents.}
  \label{fig:mmqa_supno}
\end{figure}

\begin{table*}
\begin{adjustbox}{width=1.3\columnwidth,center}
\setlength\tabcolsep{6pt}
\centering
  \begin{tabular}{@{}p{5.5cm}  cc cc cc @{}}
    \toprule
    \multirow{2}{*}{\textbf{Methods} } &  \multicolumn{2}{c }{\textbf{Single-hop}} & \multicolumn{2}{ c}{\textbf{Multi-hop}} & \multicolumn{2}{ c}{\textbf{Overall}}\\
     \cmidrule(lr){2-3}\cmidrule(lr){4-5}\cmidrule(lr){6-7}
    & F1 & EM & F1 & EM   & F1 & EM  \\
    \midrule
    \modelname{} & \textbf{75.3} & \textbf{65.7} & \textbf{70.4} & \textbf{57.6} & \textbf{73.7} & \textbf{63.9}\\
    \quad \textit{w/o graph-guided planning} & 72.4 & 61.2 & 58.6 & 45.1 & 65.3 & 53.0\\
    \quad \textit{w/o triplet conversion}  & 68.1 & 56.1 & 57.6 & 46.3 & 63.9 & 52.2\\
   \quad \textit{w/o retrieval module}  &54.0 & 44.9 & 45.8 & 33.3 & 49.8 & 39.1\\
  \bottomrule
\end{tabular}
\end{adjustbox}
\caption{The ablation studies on graph-guided planning and retrieval module. We present the F1 and EM score for single-hop, multi-hop and overall questions.}
  \label{tab:mmqa_ablation}
\end{table*}

It is likely due to the the inherent challenges of image reasoning. Without fine-tuning, vision-language models may exhibit greater variability in performance, affecting their ability to precisely match reference answers. Despite this, our approach maintains comparable performance of existing fine-tuned model.

\begin{figure*}[h]
  \centering
  \includegraphics[width=0.9\linewidth]{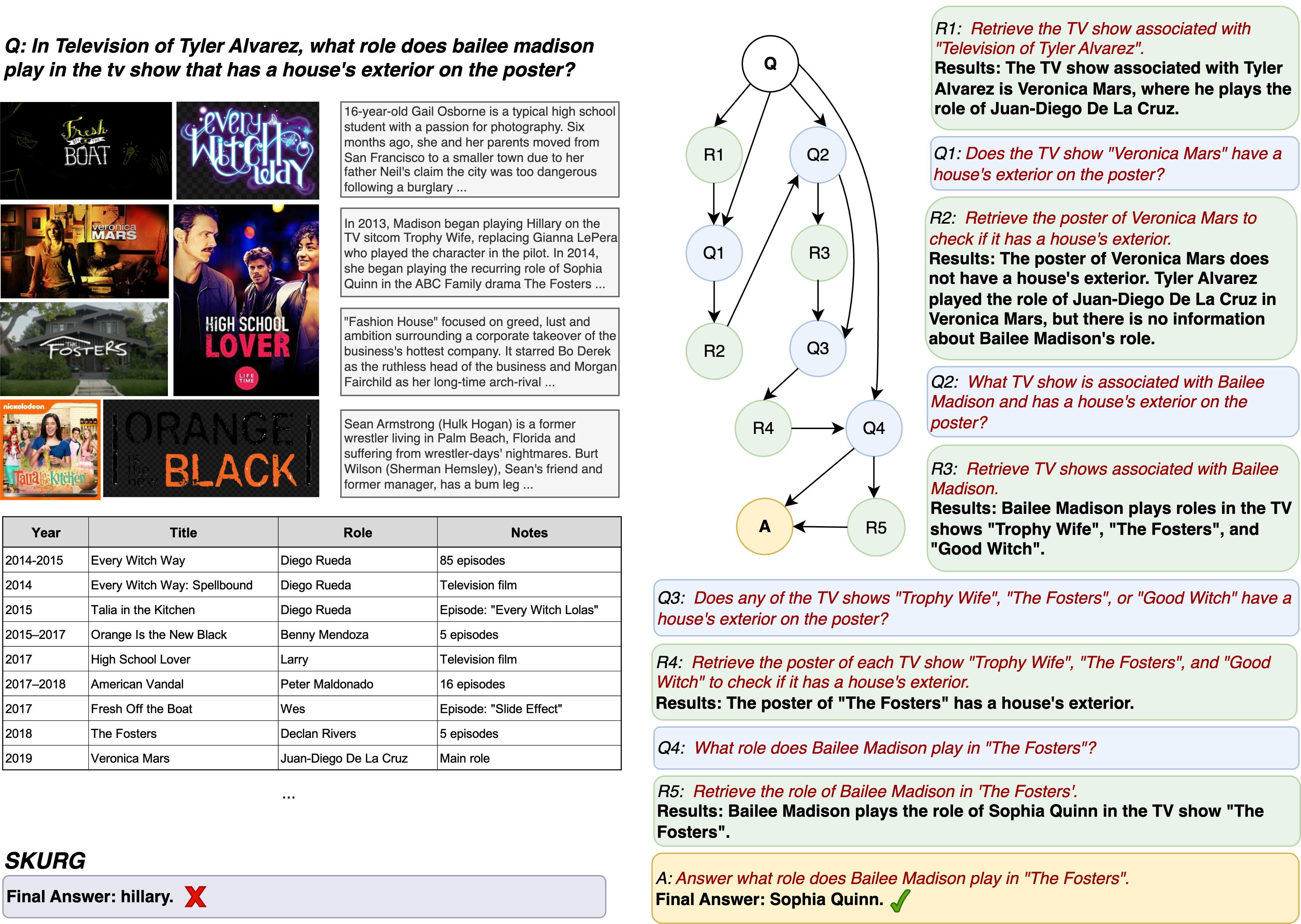}
  \caption{Case study of a multimodal multi-hop QA where \modelname{} answers correctly while SKURG fails. The question, retrieval and answer nodes are presented in blue, green and yellow boxes respectively. The instruction is displayed in red. The SKURG result is shown in a purple box.}
  \label{fig:case}
\end{figure*}

Table~\ref{tab:webqaoverall} presents the results of WebQA dataset. We observe that our method achieves accuracy comparable to that of other fine-tuned approaches. However, in terms of fluency, our method yields lower scores. Similar to the exact match scores, without fine-tuning, the paraphrased answers struggle to closely match the ground truth.

To assess our method in long-range reasoning, we compare its performance with an iterative framework, SKURG in Figure~\ref{fig:mmqa_supno}, which shows the F1 score across varying numbers of required supporting contexts in MultimodalQA. 
 As the number of supporting contexts increases, requiring more steps, SKURG's performance drops significantly, especially beyond five contexts. In contrast, our model maintains stable performance, which shows its robustness in complex reasoning scenarios.

\subsection{Ablation Study}
We present the ablation studies for \graphname{}-guided planning , triplet conversion and retrieval module on MultimodalQA.
For single-hop tasks, removing graph-guided mechanism results in minor performance decline.
However, for multi-hop questions, it leads to a substantial drop of over 10 points in both F1 and exact match scores. It demonstrates that our planning module has a critical role in handling questions that require long-range inference steps.
\add{
Eliminating triplet conversion during knowledge base construction results in consistent drops across all metrics, demonstrating its role in aligning data granularity.
}
For the retrieval module, we observe a more significant performance drop without our modality-specific strategies, as incorrect retrieval intensifies hallucination for both reasoning and global planning.
It validates that our retrieval module improve the performance even without additional training.

\subsection{Case Study}
We show a case study in Figure~\ref{fig:case}, 
which presents how the \graphname{} is constructed. 
At the initial few steps, the system attempted to retrieve information about \textit{Television of Tyler Alvarez}. However, it incorrectly shifted the focus to the TV show \textit{Veronica Mars}. Upon recognizing that this show did not align with the poster description in the question, the system redirected its reasoning toward \textbf{Bailee Madison} and relevant TV shows about the poster. 
The case study highlights that, even when the graph temporarily deviates, the system is capable of recovering and producing accurate results.

\section{Conclusion}
In this paper, we introduce a \graphname{}-guided framework for multimodal multi-hop QA. It comprises planning, retrieval, and reasoning modules, which leverage off-the-shelf models without fine-tuning. 
Compared to existing approaches, the proposed method enables flexible reasoning path exploration and plug-and-play model integration.
Experimental results show that our method achieves competitive performance against fine-tuned models even without additional training. 

\section{Limitations}
Despite the effectiveness of \modelname{}, the absence of fine-tuning leads to lower exact match and fluency scores. Additionally, frequent model calls increases inference costs. The flexibility of our  \graphname{} leads to longer exploration times and additional steps.
Future work will focus on improving module efficiency and reducing computational costs during inference.

\section{Acknowledgment}
This research was supported by the CSIRO Tech4HSE project RG242446.
Xiaoyang Wang is supported by the Australian Research Council DP230101445 and DP240101322.

\bibliography{main}

\appendix

\section{Appendices}

\subsection{Challenges of Modality-Agnostic Retrieval}

\label{appd:retrieval_challenges}

\begin{figure}[h]
  \centering
  \includegraphics[width=0.9\linewidth]{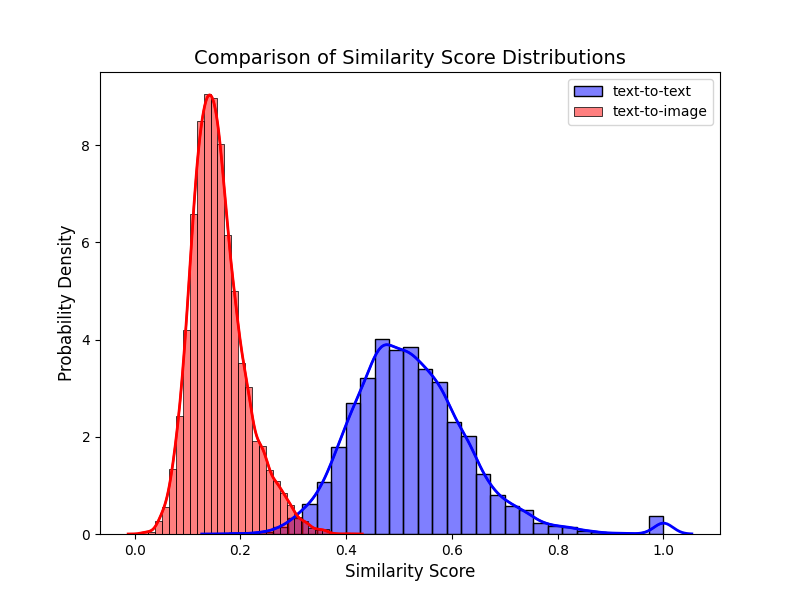}
  \caption{Comparison of similarity score distributions.}
  \label{fig:similarity_distribution}
\end{figure}

\begin{figure}[h]
  \centering
  \includegraphics[width=0.9\linewidth]{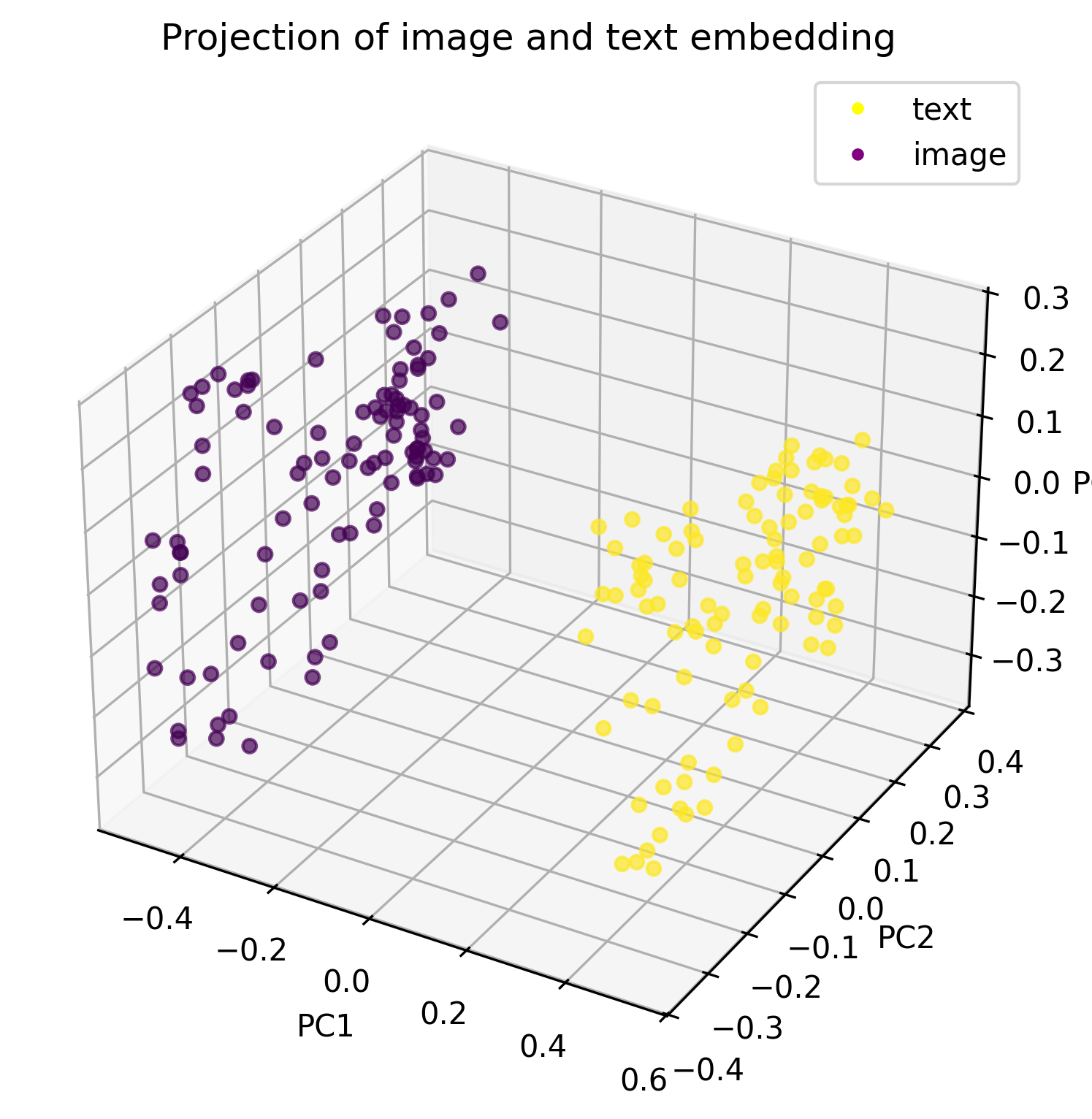}
  \caption{Projection of image and text embeddings.}
  \label{fig:modality_embed_plot}
\end{figure}

In this section, we examine the limitations of modality-agnostic retrieval when using textual queries, using 100 random selected image-caption pairs from the MSCOCO dataset. Figure~\ref{fig:modality_embed_plot} shows a 3D projection of CLIP embeddings for both images and their corresponding captions. The embeddings exhibit a clear separation by modality, which highlights the modality gap in the shared representation space. 
Furthermore, Figure~\ref{fig:similarity_distribution} presents the similarity distributions for text-text and text-image pairs derived from same set of samples, which reveals a distribution shift between text-text and text-image similarity scores. 

\subsection{Overall Workflow} \label{appd:alg}
We introduce the overall workflow in Algorithm~\ref{alg:framework}. For simplicity, we denote the prompts used for planning, retrieval and reasoning modules as $P_{planning}$, $P_{retrieval}$ and $P_{reason}$. 
Initially, separate knowledge bases are constructed for text and image modalities (line 1), and a graph is initialized with the given question (line 2).  
Following this, the graph construction procedure begins (line 3-22). At each step, the planning module provides several instructions for the generation of new node (line 4).
Depending on the new node type $\mathcal{C}_i$, the system decides which module to invoke and executes the action (Line 7-17). During this step, the content of the new node is determined and the edges are added to update the \graphname{}. 
This process continues until either a stopping condition is met or the maximum number of turns $k$ is reached. Once the graph construction phase is completed, if the final node is not of the answer type, an answer will be generated based on the current state of the graph (line 24). Otherwise, the last answer node will be returned as the final output (line 26).

\subsection{Detailed Prompts} \label{appd:prompt}
We include the prompts from the planning module mentioned in Section~\ref{sec:planning} in Table~\ref{tab:prompt}. Since the overall plan $\mathcal{P}_{plan}$ is generated based on the given question, we provide the prompt for generating the plan as $\mathcal{P}_{plan\_gen}$ instead.
The prompts from the retrieval module in Section~\ref{sec:retrieval} are listed in Table~\ref{tab:prompt_retri}. 

\add{
The few-shot examples of $\mathcal{F}_{tgt}$ and $\mathcal{F}_{descr}$ in Section~\ref{sec:retrieval} are displayed in Table~\ref{tab:few_shot_images} and Table~\ref{tab:few_shot_texts}.
We provide Example 6 as a case to deal with misclassification. For example, if $\mathcal{I}^{img}_{instr}$ corresponds to descriptive image retrieval and does not contain a specific target, after inputting into Eq.7, its $\mathcal{Q}_{text}'$ should be returned as an empty list.
}

\begin{algorithm*}[tp]
\caption{Overall workflow of the proposed framework}\label{alg:framework}
\begin{algorithmic}[1] 
\Require Question $Q$, Sources $\{S_1, S_2, \cdots, S_n\}$, A set of prompt templates $P$, max\_iteration $k$.
\State $KB_{text}, KB_{img} \gets \text{\textit{KnowledgeBaseConstruction}}(\{S_1, S_2, \dots, S_n\})$
\State $V\gets \{ Q\}$, $E \gets \{\} $, $G \gets (V, E)$ \Comment{Initialize the \graphname{}}
\For{$i = 0$ to $k$}
    \State $\mathcal{I}_{instr}, \mathcal{I}_{\mathcal{C}}, \mathcal{I}_{parent}  \gets \text{\textit{PlanningModule}}(G, P_{planning})  $ \Comment{Invoke Planning Module}
    \State $\mathcal{C}_i \gets Decompose(\mathcal{I}_{\mathcal{C}})$
    \State $v_{p1}, \cdots, v_{p_l} \gets \mathcal{I}_{parent}$ 
    \If{$ \mathcal{C}_i $ is Question}
        \State $V \gets V \cup \{v_i \}$
    \ElsIf{$ \mathcal{C}_i $ is Answer}
        \State $v_i \gets \text{\textit{ReasoningModule}} (P_{reason} ,\mathcal{I}_{instr} , \mathcal{I}_{parent})$ 
        \Comment{Invoke Reasoning module}
        \State $V \gets V \cup \{v_i \}$
    \ElsIf{$ \mathcal{C}_i $ is Retrieval}
        \State $v_i \gets \text{\textit{RetrievalModule}}(\mathcal{I}_{instr}, \mathcal{I}_{parent}, KB_{text}, KB_{img}, P_{retrieval})$ 
        \Comment{Invoke Retrieval Module}
        \State $V \gets V \cup \{v_i \}$
    \ElsIf{$ \mathcal{C}_i $ is Stop}
        \State \textbf{break} \Comment{Terminate the process}
    \EndIf
    \For{$j = 1$ to  $l$ }
        \State $E\gets E \cup \{(v_{pj}, v_i)\}$
    \EndFor
    \State Update the \graphname{} $G$ from $(V, E)$
\EndFor
\If{the type of last node $\mathcal{C}_k$ is not Answer}
    \State $A \gets \text{\textit{ReasoningModule}}( P_{reason} ,\mathcal{I}_{instr} , \mathcal{I}_{parent}, G )$
\Else
    \State $A \gets $ the content of the last answer node
\EndIf
\State \Return $A$
\end{algorithmic}
\end{algorithm*}

\subsection{Time Complexity}
\add{
In this section, we analyze the time complexity of our framework. Here we denote the time for a single LLM operation be $t_L$, a VLM operation be $t_V$, a text embedding computation be $t_{TE}$, a image embedding computation be $t_{IE}$. 
Assume there are $n$ sources. During knowledge based construction, since we utilize \textit{sklearn.BallTree} to establish the search space, the complexity will be:
\begin{align}
    O(n\cdot (t_{TE} + t_{IE})) + O(n\log n).
\end{align}
}
\add{
Assume $q$ queries are generated, a single retrieval step takes 
\begin{align}
    t_L + O(q\cdot\log n) + O(n\cdot(t_L + t_V)).
\end{align}
A reasoning step includes a single LLM operation, therefore it takes $t_L$. 
}

\add{
Let the total number of nodes be $c$, including $c_{retr}$ retrieval steps and $c_{reason}$ reasoning steps. The planning stages would take $c \cdot t_L$ in total. 
The overall time complexity is displayed as following: 
\begin{align}
\begin{split}
  &O(n\cdot (t_{TE} + t_{IE})) + O(n\log n) \\
  &+ c_{retr} \cdot (t_L + 
  O(q\cdot\log n) 
  + O(n\cdot(t_L + t_V))) \\  
  &+ c_{reason}  \cdot t_L + c \cdot t_L.  
  \end{split} 
\end{align}
}

\begin{figure}
  \centering
  \includegraphics[width=0.9\linewidth]{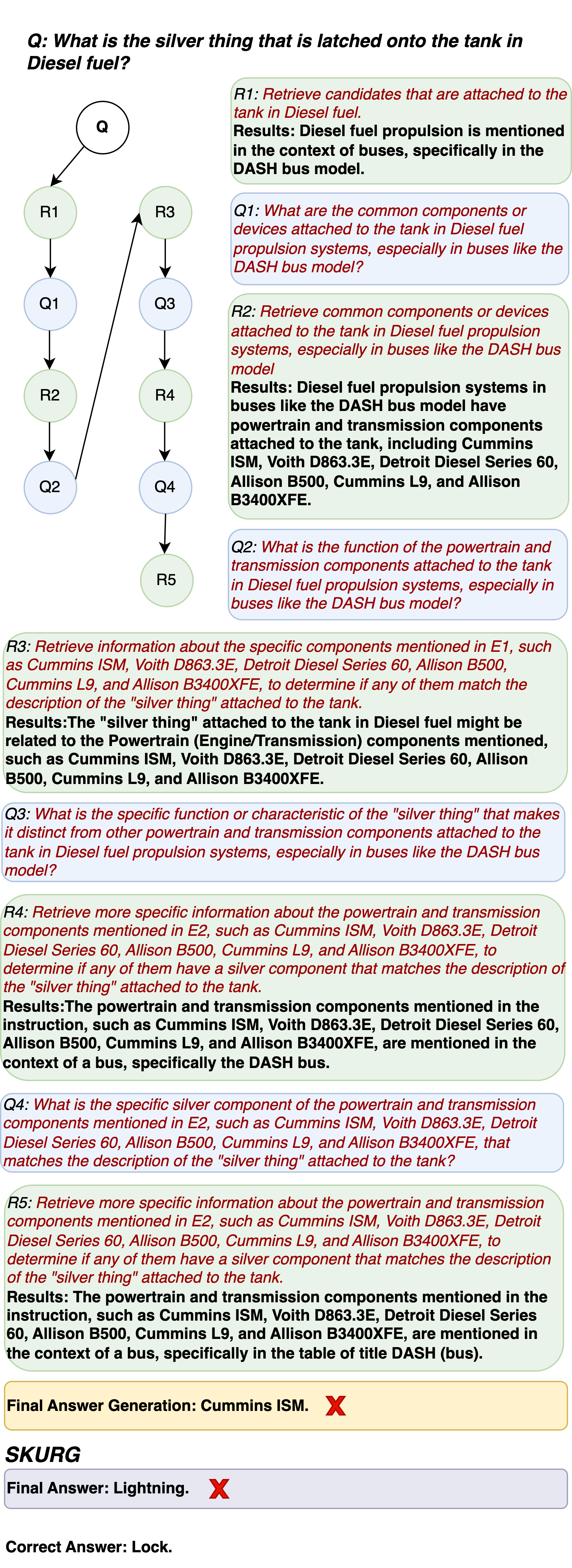}
  \caption{Comparison of similarity score distributions}
  \label{fig:bad_case_study}
\end{figure}

\subsection{Implementation } \label{appd:impl}
In our training-free framework, models can be easily incorporated as required, without the need for fine-tuning. Here, we utilize \textit{Llama3-70B-Instruct} as the LLM and \textit{LLaVA-13B} as the visual-language model. As outlined in Section~\ref{sec:kb_construction}, the text knowledge base supports search within a text embedding space, while the image knowledge base expects cross-modal retrieval between text queries and images. To facilitate these retrieval processes, we employ CLIP to generate embeddings for both texts and images. 
\add{
During knowledge base construction, we utilize \textit{sklearn.BallTree} structure to efficiently store and organize the embeddings.
To retrieve candidates, we employ \textit{query$\_$radius} method, which returns all neighbors within a predefined distance threshold. 
We choose this method over top-$k$ retrieval as top-$k$ assumes a fixed number of relevant items, which is unsuitable for multimodal multi-hop question answering tasks, where the number of relevant sources can vary depending on the question.
}

For datasets that include the table modality, we preprocess tables by converting the structured data from each row into a sentence. They are then treated similarly to text modality for subsequent processing.

\begin{table*}
\begin{adjustbox}{width=1.8\columnwidth,center}
\setlength\tabcolsep{10pt}
\renewcommand{\arraystretch}{1.3} 
  \begin{tabular}{m{0.2\linewidth} | p{0.8\linewidth} }
    \hline
    \textbf{Category} & \textbf{Description} \\
    \hline
    \hline
    \textbf{Incorrect} & The prediction does not match with ground truth answer at all. \\
    \hline
 \textbf{Mostly Correct} &  The prediction has the same meaning as the ground truth answer but receives a zero EM score due to minor discrepancies such as number formatting, date representation, or the use of symbols.\\
    \hline
    \textbf{Incomplete} & The prediction provides a partial response, failing to fully capture the complete content of the ground truth answer.  \\ 
    \hline
    \textbf{Abbreviation} & The prediction reflects the same meaning as the ground truth answer but is presented in an abbreviated form.\\
    \hline
    \textbf{Overlap} & The prediction partially aligns with the ground truth, sharing some overlapping content but lacking a complete match. \\
    \hline
    \textbf{Redundant} & The prediction includes the complete ground truth answer but is characterized by the presence of extraneous or redundant components. \\
    \hline
\end{tabular}
\end{adjustbox}
\caption{Descriptions of the categories in error analysis.}
  \label{tab:error_reason}
\end{table*}

\begin{figure}
  \centering
  \includegraphics[width=0.9\linewidth]{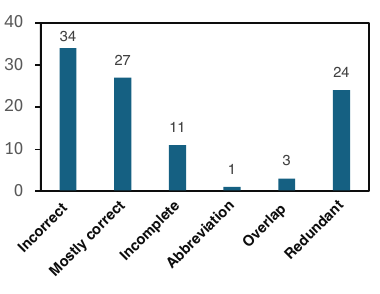}
  \caption{The error analysis of cases where the EM score is zero.}
  \label{fig:mmqa_em_analyze}
\end{figure}

\subsection{Dataset Statistics}
In this section, we present the statistics of the datasets used. 
WebQA dataset comprises 34K training samples, 5K development samples and 7.5K test samples. MultimodalQA dataset includes 23K training samples, 2.4K development samples and 3.6K test samples.

\subsection{Ablation Study Setup} \label{appd:abl_setup}
Firstly, to evaluate the performance without planning module, we modify the framework by omitting the \graphname{} construction. In this setting, the actions are planned in a sequential manner, where each step is built directly upon previous one, similar to an iterative framework. Consequently, the system is constrained to a single-path approach, alternating between retrieval and reasoning without the flexibility to explore other paths. 
Secondly, we evaluated the performance without our retrieval module by replacing it with a simplified version. It retrieves information based solely on similarity score between the embeddings of instruction queries and sources. The key components of our retrieval design, including query construction and candidate examination, are eliminated.

\subsection{Case Study}
\add{
We demonstrate an example for failure cases in Figure~\ref{fig:bad_case_study}.
The potential failure points arise because nodes continue providing seemingly relevant information at starting point, leading the system to follow this path even though it does not capture the truely relevant sources. It repeatedly extracts the same context, mistakenly believing it may contain the necessary details. Additionally, the graph state summary gradually becomes too long, making it harder for the LLM to produce the correct action, as handling long contexts remains a challenge for LLMs. The complexity of such questions also applies to the baseline method~\cite{yang2023enhancing}, highlighting the inherent challenges posed by indistinguishable distractors.}



\begin{table*}
\begin{adjustbox}{width=1.8\columnwidth,center}
\setlength\tabcolsep{10pt}
\renewcommand{\arraystretch}{1.3} 
  \begin{tabular}{ | m{0.1\linewidth} | p{0.8\linewidth} |}
    \hline
    \textbf{Notation} & \textbf{Prompt Template} \\
    \hline
    $\mathcal{P}_{plan\_gen}$ & Given a multi-hop question, break the question into key components to identify all necessary information. \\
    \hline
    $\mathcal{P}_{parent}$ &  Specify which existing node(s) will be used as the foundation for generating a new node, ensuring a clear and logical progression in the reasoning process.\\
    \hline
    $\mathcal{P}_{\mathcal{C}}$ & Please analyze the content of existing nodes step by step, and then decide what new node to generate. Here are some options:
    \begin{enumerate}
        \item Stop: when the answer to question Q is already found given the current nodes.
        \item Retrieval: to retrieve candidates and extract useful information from candidates based on existing nodes.
        \item Answer: to produce the instruction to generate an answer based on an existing question and other nodes.
    \end{enumerate}
    \\
    \hline
    $\mathcal{P}_{state}(G)$ & 
    Here is the graph: ...

    Your task is to determine the next step in deriving the final answer of $Q$ based on provided input. Please think step by step and consider the current reasoning graph and overall plan.

    \\
    \hline
\end{tabular}
\end{adjustbox}
\caption{Prompts for Planning Module.}
  \label{tab:prompt}
\end{table*}

\begin{table*}
\begin{adjustbox}{width=1.8\columnwidth,center}
\setlength\tabcolsep{10pt}
\renewcommand{\arraystretch}{1.3} 
  \begin{tabular}{ | m{0.1\linewidth} | p{0.8\linewidth} |}
    \hline
    \textbf{Notation} & \textbf{Prompt Template} \\
    \hline
    $\mathcal{P}_{decomp}$ & Given the following instruction, your task is to identify and extract the text-related part and the image-related part for retrieval. The instruction may contain references to both textual and visual content. For image-related parts, determine if the task is Targeted Image Retrieval (specific images named) or Descriptive Image Retrieval (search based on description).\\
    \hline 
    $\mathcal{P}_{extract}$ & Given the following text, your task is to extract the keywords. The keywords should be specific and helpful for retrieval. \\
    \hline
    $\mathcal{P}_{tgt}$ & For targeted image retrieval, please list specific image if mentioned, and craft a precise question based on the image-related request to guide the assistant on what to identify or analyze in the image.\\
    \hline
    $\mathcal{P}_{descr}$ & For descriptive image retrieval, please extract descriptive phrase, and craft a precise question based on the image-related request to guide the assistant on what to identify or analyze in the image.\\
    \hline
    $\mathcal{P}_{exam}^{text}$ & Your task is to analyze the input text and check if it is related to the instruction: ...\\
    \hline
    $\mathcal{P}_{exam}^{img}$ & Given the image, provide a brief description of its content and answer the question based on the provided instruction: ...\\
    \hline
    $\mathcal{P}_{retr}$ & These are the instruction and retrieval results: ...

    Please extract only the relevant information from the result and rephrase into valid description as the corresponding answer to the instruction for later analysis. 
    \\
    \hline
\end{tabular}
\end{adjustbox}
\caption{Prompts for Retrieval Module.}
  \label{tab:prompt_retri}
\end{table*}

\begin{table*}
\begin{adjustbox}{width=1.8\columnwidth,center}
\setlength\tabcolsep{10pt}
\renewcommand{\arraystretch}{1.3} 
  \begin{tabular}{ | m{0.1\linewidth} | p{0.8\linewidth} |}
    \hline
    \textbf{Notation} & \textbf{Few-shot Examples} \\
    \hline
    $\mathcal{F}_{text}$ & 
    Example 1: 
    
Instruction: Retrieve the NHL team played against the Pittsburgh Penguins in the playoff series.

Key Phrase:["NHL team played against the Pittsburgh Penguins in the playoff series"]

Example 2: 

Instruction:  Retrieve the 1977 Seattle Seahawks Kingdome regular season opponent that has the most Super Bowl losses in NFL history.

Key Phrase: ["1977 Seattle Seahawks Kingdome opponent", "Super Bowl losses in NFL history"]

Example 3: 

Instruction: Retrieve the role Peppe Lanzetta played in the 2009 film.

Key Phrase: ["role Peppe Lanzetta played in the 2009 film"]

Example 4:  

Instruction: Retrieve the Magazine that had Caroline Miller as Editor in Chief and the year it won a National Magazine Award.

Key Phrase: ["Magazine that had Caroline Miller as Editor", "year of Magazine won a National Magazine Award"]

Example 5: 

Instruction: Retrieve the song performed in episode 3 of season 1 of The Clash on July 14.

Key Phrase: ["song performed in The Clash on July 14", "episode 3 of season 1 of The Clash on July 14"].

   \\
    \hline 
\end{tabular}
\end{adjustbox}
\caption{Few-shot examples for $\mathcal{F}_{text}$.}
  \label{tab:few_shot_texts}
\end{table*}

\begin{table*}
\begin{adjustbox}{width=1.8\columnwidth,center}
\setlength\tabcolsep{10pt}
\renewcommand{\arraystretch}{1.3} 
  \begin{tabular}{ | m{0.1\linewidth} | p{0.8\linewidth} |}
    \hline
    \textbf{Notation} & \textbf{Few-shot Examples} \\
   
    \hline

   $\mathcal{F}_{tgt}$ & Example 1:

   Instruction: Retrieve the structure at the top of the Stockport County F.C.'s logo.

   Question: What is the structure at the top of logo?

   Target: ["the Stockport County F.C.'s logo"]

    Example 2:

   Instruction: Retrieve the colors that make up the flag for Denmark.

   Question: What are the colors that make up the flag?

   Target: ["flag for Denmark"]

   Example 3:

   Instruction: Retrieve the number of leaves are on the clover of the Celtic F.C.'s logo.
   
   Question: How many leaves are on the clover of the logo?

   Target: ["Celtic F.C.'s logo"]

   Example 4:

   Instruction: Retrieve the hair style of the man in blue in American football.
   
   Question: What hair style does the man in blue have in American football?

   Target: ["American football"]

   Example 5:

   Instruction: Retrieve whether there is a fence around the outside of the Cotton Bowl (stadium).
   
   Question: Is there a fence around the outside of the Cotton Bowl?

   Target: ["the Cotton Bowl (stadium)"]

    Example 6:

   Instruction: Retrieve the movie poster that has a woman with a green dress on it.
   
   Question: Is there a woman with a green dress on it?

   Target: []

   \\
\hline
   
   $\mathcal{F}_{descr}$ & Example 1:

   Instruction: Retrieve the team whose logo has a bird on it.

   Question: Is there a bird in the logo?

   Key Phrase: ["logo with a bird"]

    Example 2:

   Instruction: Retrieve the movie poster that has a woman with a green dress on it.

   Question: Is there a woman with a green dress on it?

   Key Phrase: ["movie poster that has a woman with a green dress"]

   Example 3:

   Instruction: Retrieve the television title with a car on its poster.
   
   Question: Is there a car on the poster?

   Key Phrase: ["television poster with a car on it"]

   Example 4:

   Instruction: Retrieve opponent that has a football helmet on its logo.
   
   Question: Is there a football helmet on the logo?

   Key Phrase: ["logo with a football helmet"]

   Example 5:

   Instruction: Retrieve the poster that has a man reaching out with his hand.
   
   Question: Is there a man reaching out with his hand?

   Key Phrase: ["poster with a man reaching out with his hand."]

    Example 6:

   Instruction: Retrieve the number of leaves are on the clover of the Celtic F.C.'s logo.
   
   Question: How many leaves are on the clover of the logo?

   Key Phrase: []

   \\
    \hline 
\end{tabular}
\end{adjustbox}
\caption{Few-shot examples for $\mathcal{F}_{tgt}$ and $\mathcal{F}_{descr}$.}
  \label{tab:few_shot_images}
\end{table*}

\subsection{Error analysis} \label{appd:err}

To gain a deeper understanding of the causes of errors, we conduct an error analysis by randomly sampling instances from the results where the exact match (EM) score is zero. We categorize the reasons for these errors into six distinct categories, with detailed descriptions provided in Table~\ref{tab:error_reason}. As depicted in Figure~\ref{fig:mmqa_em_analyze}, we observe that 27$\%$ of the results, despite being mostly correct, were classified as inaccurate due to the strict criteria of the exact match metric. Another significant source of error was the presence of redundant results, which generate more outputs than expected.

\end{document}